\documentclass[conference]{IEEEtran}
\usepackage{cite}
\usepackage{amsmath,amssymb,amsfonts}
\usepackage{algorithmic}
\usepackage{graphicx}
\usepackage{textcomp}
\usepackage{xcolor}

\usepackage{subcaption}
\usepackage{tikz}
\usetikzlibrary{positioning, calc, shapes.geometric}

\def\BibTeX{{\rm B\kern-.05em{\sc i\kern-.025em b}\kern-.08em
    T\kern-.1667em\lower.7ex\hbox{E}\kern-.125emX}}
\begin{document}

\newcommand\copyrighttext{%
  \footnotesize \textcopyright 2023 IEEE. Personal use of this material is permitted.
  Permission from IEEE must be obtained for all other uses, in any current or future
  media, including reprinting/republishing this material for advertising or promotional
  purposes, creating new collective works, for resale or redistribution to servers or
  lists, or reuse of any copyrighted component of this work in other works.}
\newcommand\copyrightnotice{%
\begin{tikzpicture}[remember picture,overlay]
\node[anchor=south,yshift=10pt] at (current page.south) 
  {\fbox{\parbox{\dimexpr\textwidth-\fboxsep-\fboxrule\relax}{\copyrighttext}}};
\end{tikzpicture}%
}

\title{Connections Between Pairs of Filters Improve the Accuracy of Convolutional Neural Networks\\
}
\author{\IEEEauthorblockN{1\textsuperscript{st} Kathleen Anderson}
\IEEEauthorblockA{\textit{Institute for Neuro- and Bioinformatics} \\
\textit{University of Lübeck}\\
Germany \\
k.anderson@uni-luebeck.de}
\and
\IEEEauthorblockN{2\textsuperscript{nd} Philipp Grüning}
\IEEEauthorblockA{\textit{Institute for Neuro- and Bioinformatics} \\
\textit{University of Lübeck}\\
Germany \\
ph.gruening@uni-luebeck.de}
\and
\IEEEauthorblockN{3\textsuperscript{rd} Erhardt Barth}
\IEEEauthorblockA{\textit{Institute for Neuro- and Bioinformatics} \\
\textit{University of Lübeck}\\
Germany \\
erhardt.barth@uni-luebeck.de}
}

\maketitle
\copyrightnotice

\begin{abstract}

While researchers continue to find new and improved network structures for CNNs, most of the newly invented architectures still rely on the traditional pattern of stacking convolutional blocks and separating them with pointwise activation functions. However, there are drawbacks to a network purely building on pointwise nonlinearities. One alternative is to introduce a pairwise connection between two filters of a network. Typical connection functions use multiplications or the minimum operation to realize logical AND connections.
In this paper, we go one step further by demonstrating that CNNs can benefit from more general connections, which include parameters that are learned. With such parameters, the network is able to implement different connections in different network layers and better adapt the connection function to the task at hand.

\end{abstract}


\section{Introduction} 
Convolutional neural networks (CNNs) are one of the most powerful solutions for many computer vision problems \cite{li2021survey,Sandler2018MobileNetV2,He2016ResNets, Krizhevsky2017AlexNet,Szegedy2016InceptionV4}. The field of research concerning CNNs is vast and involves various ideas to improve them. Some methods add more and more layers, introducing new ideas to help larger models generalize \cite{szegedy2015inception,He2016ResNets}. Others help the training process and make it easier to find the optimal weights (e.g. batch normalization \cite{Ioffe2015BatchNorm} and dropout \cite{Hinton2012Dropout}). 

The methods above usually rely on the conventional structure: a convolution followed by a pointwise nonlinearity. As pointed out in \cite{paiton2020PopulationNonlinearity}, a network defined by simple pointwise nonlinearities is generally not selective enough, and thus not robust against adversarial attacks. The authors present population nonlinearities as a better alternative: nonlinearities where the output is a function of more than one neuron. In that same regard, bilinear, Volterra, and polynomial approaches \cite{li2017factorizedBilinear,zoumpourlis2017quadraticInteraction,Chrysos2020PNets} are one way of implicitly connecting pairs of neurons.

Feature-Product (FP) networks \cite{gruning2020FPYieldEffecientNetworks,gruning2020log,gruning2021fpImageQualAssess,gruning2022MinNets,gruning2022FPInspiredByVision} contain FP-blocks which implement a pairwise interaction by combining the output of two convolutions in an AND fashion, aiming to filter the input based on its intrinsic dimensionality \cite{barth2000unique2D}. The intrinsic dimension relates to the local image structure: dimension 0 indicates an area with no change in intensity, a one-dimensional (1D) area refers to an area varying in only one direction (e.g. straight edges and lines), and a two-dimensional (2D) area varies in two directions (e.g. corners and junctions). 

FP-nets are inspired by human vision: in certain areas of the human brain, end-stopped cells are thought to detect image areas with intrinsic dimension 2 \cite{hubelEndStopping1965}. 

Based on FP-blocks, which realize an AND-connection via multiplication, Min-nets~\cite{gruning2022MinNets} showed that the minimum operation is also suitable.
The associated \textit{Min block} is made up of two independently trained depthwise convolutions \cite{Chollet2017XceptionDL} with subsequent ReLU ($+$). For each pixel position, the respective values $x^+$ and $y^+$ of the two feature maps are combined using $\min(x^+, y^+)$.

By AND combining the outputs of two filters, Min blocks have the capacity to detect 2D areas in an input image. If one filter reacts to variation in one direction and the other responds to a different direction, the product of both filter maps highlights areas varying in two directions - 2D areas. Those areas are unique and sparse in natural images\cite{barth2000geometric}, and the information contained in them suffices to recreate the image \cite{barth2000unique2D}. Therefore, a 2D filter can generate an efficient representation of the input. 

However, while the 2D filter provides an advantage on natural images, the idea of extracting only the corners or junctions could be too restrictive in deeper layers of the network where the feature map is no longer resembling a natural image.

A typical AND block combines the outputs of two filters either with a connection function based on multiplication (FP block) \cite{gruning2022FPInspiredByVision} or a function based on the minimum operation (Min block) \cite{gruning2022MinNets}. In this paper, we keep the overall structure of an AND block, which we refer to as a \textit{connected block}. However, we show that with more general connection functions, we can achieve better results. In a connected block, an input feature map is processed with two filters. For each pixel position $(i,j)$, the connection function combines the output of filter one with the output of filter two – but that function is not limited to an AND logic. Using different connection functions, the connected block acts not only as a filter but is also capable of distinguishing different areas without forcefully removing information.

We start by showing that pairwise connections other than the previously used AND connections can improve the performance of CNNs. Subsequently, we present a novel connection function that includes a learnable parameter. The parameter defines whether the connection merges the two inputs in a rather linear way, or transforms them to highlight different areas with positive or negative output values. The improved accuracy proves that a more flexible function, which can be adjusted to different layers of the network, can be a powerful enhancement. 

Our idea of a parameterized connection function between two depthwise convolutions of a CNN combines the concept of parameterized activation functions (such as the PReLU function \cite{He2015PReLU} and the learning piecewise linear function presented in \cite{agostinelli2014learningPiecewiseLinearActivation}) with the idea of pairwise feature interactions. Bilinear CNNs \cite{Lin2015BilinearCNN, li2017factorizedBilinear} are a prominent example of the effective utilization of pairwise interactions. A similar interaction between the pixels of a feature map has been used as a Volterra-based convolution in \cite{zoumpourlis2017quadraticInteraction}. In a $\Pi$-net \cite{Chrysos2020PNets}, a skip connection similar to a residual connection is implemented as an elementwise multiplication. 

Bilinear CNNs and Volterra-based convolutions, as well as the original FP-net architecture, are built around a multiplicative relation between one or two feature maps. As seen in \cite{gruning2022MinNets,gruning2020log}, and demonstrated in the experiments of this paper, more general connection functions can increase the performance using the same number of learnable parameters.

\section{Connected Blocks} 
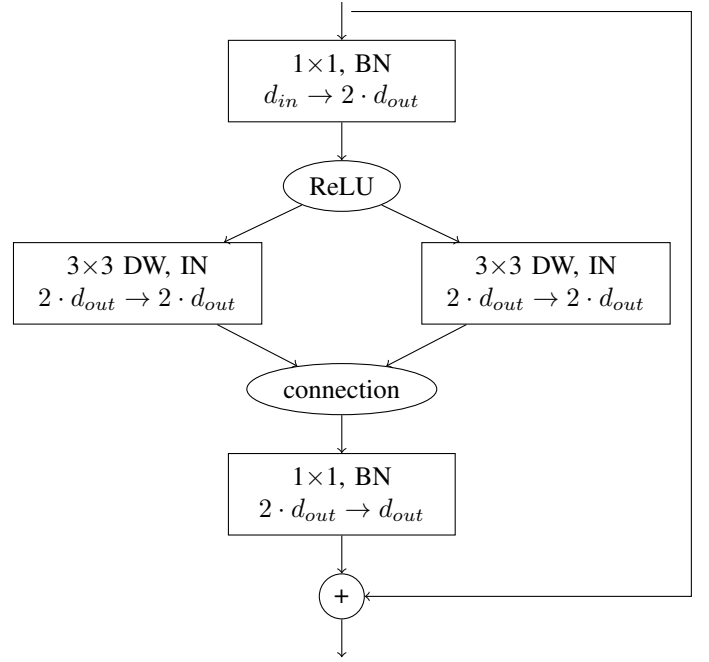
\begin{figure}
    \centering
        \begin{tikzpicture}[
            node distance=5mm and 5mm,
            block/.style={
                draw,
                minimum width={3cm},
            }]
            \node[] (b0) {};
            \node[block, below=of b0] (b1) {\begin{tabular}{c} 1$\times$1, BN \\ $d_{in} \rightarrow 2 \cdot d_{out}$\end{tabular}};
            \node[ellipse, draw, below=of b1] (b2) {ReLU};
            \node[block, below left=of b2] (b3a) {\begin{tabular}{c} 3$\times$3 DW, IN \\ $2 \cdot d_{out} \rightarrow 2 \cdot d_{out}$\end{tabular}};
            \node[block, below right=of b2] (b3b) {\begin{tabular}{c} 3$\times$3 DW, IN \\ $2 \cdot d_{out} \rightarrow 2 \cdot d_{out}$\end{tabular}};
            \node[ellipse, draw, below=2cm of b2] (b4) {connection};
            \node[block, below=of b4] (b5) {\begin{tabular}{c} 1$\times$1, BN \\ $2 \cdot d_{out} \rightarrow d_{out}$\end{tabular}};
            \node[circle, draw, below=of b5] (b6) {+};
            \node[below=of b6] (b7) {};

            \draw[->] (b0) edge (b1); 
            \draw[->] (b1) edge (b2); 
            \draw[->] (b2) edge (b3a) edge (b3b); 
            \draw[->] (b3a) edge (b4); 
            \draw[->] (b3b) edge (b4); 
            \draw[->] (b4) edge (b5); 
            \draw[->] (b5) edge (b6); 
            \draw[->] (b6) edge (b7); 

            \draw [->] ($(b0.east) + (0, -0.25)$) -| +(4.5,-0.5) |- (b6.east);
        \end{tikzpicture}
    \caption{Structure of a connected block, illustrated as the sequence of actions performed and the number of input and output channels. 3$\times$3 indicates a convolution with kernel size 3, DW stands for a depthwise convolution. BN and IN are abbreviations for batch and instance normalization, respectively. The residual connection is visualized with the "+" node at the bottom of the figure. We use zero padding in the residual connection to make the block input match the block output.}
    \label{fig:connected_block}
\end{figure}

The structure of the \textit{connected block} presented in this paper largely follows the design of AND blocks as introduced in \cite{gruning2022FPInspiredByVision}. Our implementation is depicted in Figure~\ref{fig:connected_block}: starting with a pointwise convolution, batch normalization, and ReLU activation, the block then splits into two parallel depthwise convolutions, which are normalized using instance normalization \cite{Ulyanov2016InstanceNorm} and then merged using the nonlinear connection function and another pointwise convolution. The connection function can also include ReLUs.

Given two output feature maps of the left depthwise convolution $X = (x_0,..., x_n)$ and the right depthwise convolution $Y = (y_0, ..., y_n)$ (with $n$ as the number of points in $X$ and $Y$), the combined feature map $Z = (z_0,..., z_n)$ is computed using the scalar connection function $z_i = f(x_i, y_i) \ \forall i \in \{0, ..., n\}$. The same connection function is used for each point of each feature map.

\begin{figure*}
\centering
    \begin{subfigure}{0.31\textwidth}
        \centering
        \includegraphics[width=\textwidth]{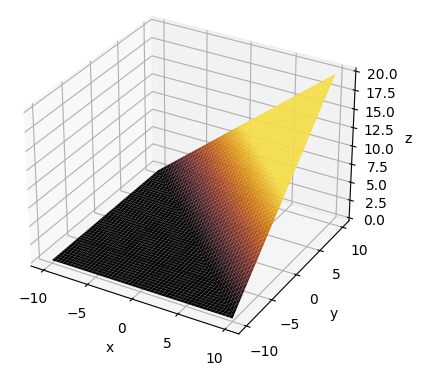}
        \caption{Simple connection \\
        $(x+y)^+$}
        \label{fig:mesh_non}
    \end{subfigure} \hspace{0.2cm}
    \begin{subfigure}{0.31\textwidth}
        \centering
        \includegraphics[width=\textwidth]{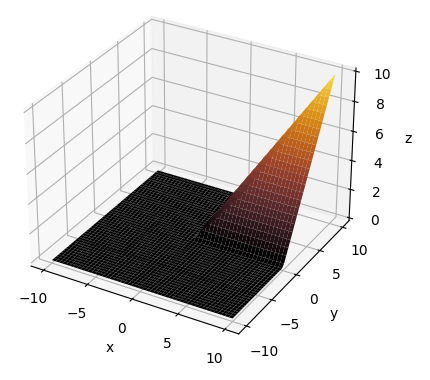}
        \caption{Minimum based AND function \\
        $\min(x^+, y^+)$}
        \label{fig:mesh_min}
    \end{subfigure}  \\
    \begin{subfigure}{0.31\textwidth}
        \centering
        \includegraphics[width=\textwidth]{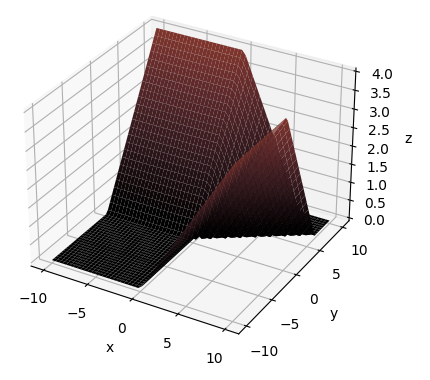}
        \caption{XOR inspired function with ReLU\\
        $\max(x^+, y^+) - 0.5 \cdot (x^+ + y^+)$}
        \label{fig:mesh_reluxor}
    \end{subfigure} \hspace{0.2cm}
    \begin{subfigure}{0.31\textwidth}
        \centering
        \includegraphics[width=\textwidth]{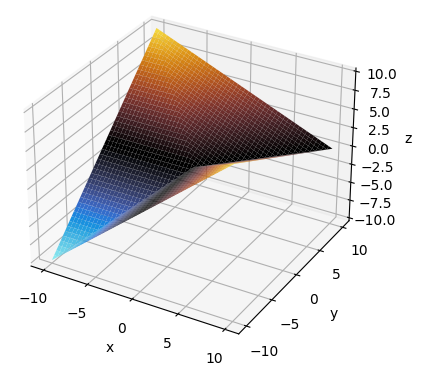}
        \caption{XOR inspired function without ReLU\\
        $\max(x, y) - 0.5 \cdot (x + y)^+$}
        \label{fig:mesh_xor}
    \end{subfigure}
    \caption{Meshgrids illustrating the four most relevant connection functions. The x- and y-axis depict the two inputs for the scalar connection function, the output of the function is drawn on the z-axis.}
    \label{fig:meshgrids_fixed}
\end{figure*}

We compare our new connection functions to two baseline models. Both baselines utilize the same block structure as depicted in Figure~\ref{fig:connected_block}, including the two parallel depthwise convolutions. Firstly, we verify that it is in fact the connection function, not the block structure, that is affecting the results. The \textit{simple} connection function, implemented as 
    \begin{equation}
        f(x, y) = (x+y)^+
        \label{eq:simple}
    \end{equation}
(note that we use $a^+=\max(a, 0)$ to describe a ReLU activation), is inspired by residual connections \cite{He2016ResNets} and the fusing process of Non-Deep Networks \cite{goyal2021nonDeepNetworks}. As seen in Figure~\ref{fig:mesh_non}, the output increases linearly with its two inputs, as long as $x+y$ is positive. 

The idea of parallel convolutions has been proven to have potential on its own \cite{goyal2021nonDeepNetworks}. However, we intend to do more than merge two parallel blocks. The connected blocks are meant to transform the output, highlight areas that contain relevant data, and even differentiate between different types of information (e.g., the intrinsic dimensionality). An accuracy below the simple connection baseline indicates that the input is not changed in a way that improves the model. 

As a second baseline, we use an AND connection as implemented in \cite{gruning2022MinNets}: 
\begin{equation}
    f(x, y) = \min(x^+, y^+).
\end{equation}
A minimum AND connection sets the $f(x, y)$ to zero if one (or both) inputs are equal to or less than zero. Only if both $x$ and $y$ are positive the output is equal to the smaller one of the input values (see Figure~\ref{fig:mesh_min}). A connected block with a minimum function (Min block) can verifiably increase the test accuracy of a network \cite{gruning2022MinNets}. But does a connected block have to realize the AND logic to achieve this result? We evaluated various different connection functions to answer that question. One of the most successful new functions is, again, inspired by a logic gate: the XOR connection.

An XOR connection can be more easily understood using its original form, a ReLU-activated XOR, using the formula:
\begin{equation}
    f(x, y) = \max(x^+, y^+) - l \cdot (x^+ + y^+).
    \label{eq:xor_relu}
\end{equation}
To realize the original XOR interpretation, we fixed the parameter $l$ to 0.5 for all connections. The function is also displayed in Figure~\ref{fig:mesh_reluxor}. Given the two input points $x$ and $y$ and w.l.o.g. $x\geq y$:
\begin{itemize}
    \item If $x \leq 0$ or $y \leq 0$, the output is zero.
    \item If $x$ is positive and $y$ negative or small, the output is equal or close to $x - 0.5 x = 0.5 x$. 
    \item If $x$ and $y$ are relatively close to each other, $\max(x, y) = x$ can no longer exceed $0.5 \cdot (x + y)$. The closer they are, the smaller the difference between $x$ and $0.5 \cdot (x + y)$, and the closer the output approaches zero. 
\end{itemize}

A variation of Equation \ref{eq:xor_relu} that generally outperformed the implementation above omits the inner ReLU activation:
\begin{equation}
    f(x, y) = \max(x, y) - l \cdot (x + y)^+.
    \label{eq:xor}
\end{equation}
The main difference, compared to the original XOR, is that $x < 0$ or $y < 0$ now causes negative outputs, entailing that $x \approx y$ can now, to a limited extent, lead to an output not equal to zero (for $l=0.5$, see Figure~\ref{fig:mesh_xor}). 

\begin{figure*}
\centering
    \begin{subfigure}{0.31\textwidth}
        \centering
        \includegraphics[width=\textwidth]{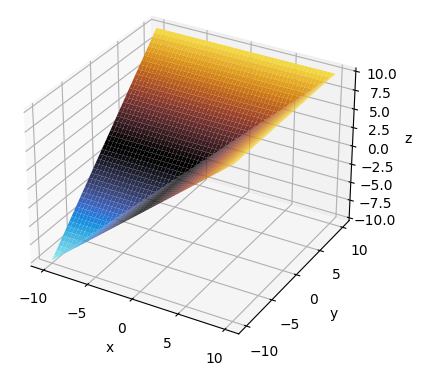}
        \includegraphics[width=\textwidth]{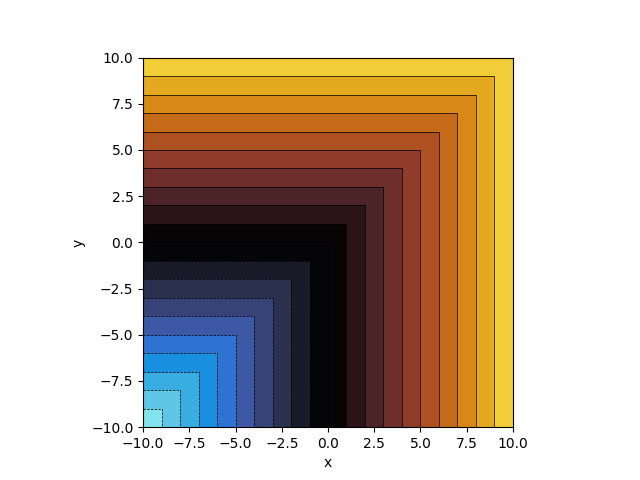}
        \caption{$l = 0$}
        \label{fig:mesh_xor_0}
    \end{subfigure} \hspace{0.2cm}
    \begin{subfigure}{0.31\textwidth}
        \centering
        \includegraphics[width=\textwidth]{images/xor_nothing_05_c.png}
        \includegraphics[width=\textwidth]{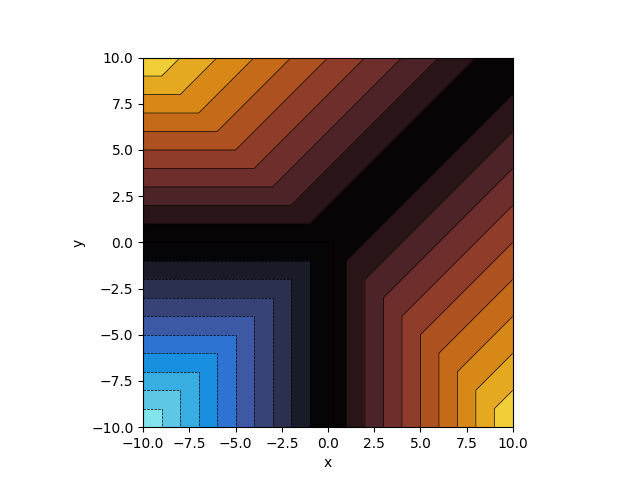}
        \caption{$l = 0.5$}
        \label{fig:mesh_xor_05}
    \end{subfigure} \hspace{0.2cm}
    \begin{subfigure}{0.31\textwidth}
        \centering
        \includegraphics[width=\textwidth]{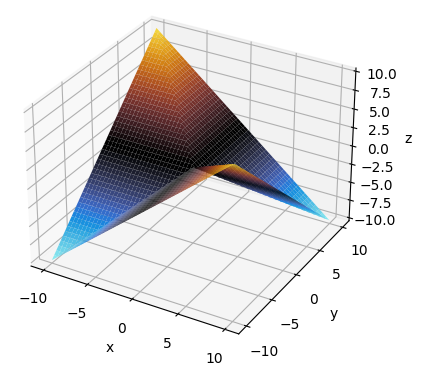}
        \includegraphics[width=\textwidth]{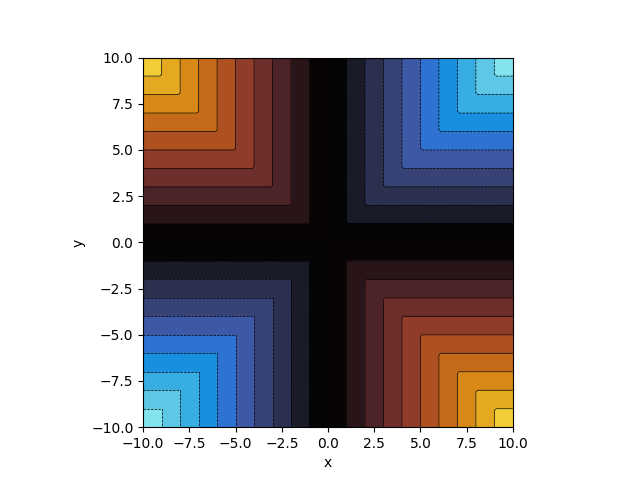}
        \caption{$l = 1$}
        \label{fig:mesh_xor_1}
    \end{subfigure} \hspace{0.2cm}
    
    \caption{Meshgrid (top row) and contour plots (bottom row) for three different connections that can be realized by the function $\max(x, y) - l \cdot (x+y)^+$. The x- and y-axis depict the two inputs for the scalar connection function ($x$ and $y$), the output of the function is drawn on the z-axis (top row) and given as the color of an area.
    Orange shows positive values and blue negative values. A lighter color indicates a higher absolute value and darker colors indicate a lower absolute value with black areas having the value 0.
    }
    \label{fig:meshgrids_learning}
\end{figure*}

The XOR-inspired function in Equation~\ref{eq:xor} includes a constant parameter ($l = 0.5$) that balances the two terms of the function. In simple terms, for the output to be positive, the greater of the two input values ($\max(x, y)$) has to exceed the mean of both ($0.5 \cdot (x + y)^+$). However, when relaxing the strict XOR interpretation, said parameter can shift the function to express different concepts. For the experiments presented in this paper, we limited $l$ to $[0, 1]$.
\begin{itemize}
    \item When $l=0$, the second term is eliminated, leaving the connection function $\max(x, y)$ (also seen in Figure~\ref{fig:mesh_xor_0}). The connection can be interpreted as an OR, a simple combination of the two sides: when one or both convolutions highlight a certain area, that area will be highlighted with the same activation values in the combined output. 
    \item On the other hand, if $l \gg 0.5$, the second term (the combination of $x$ and $y$) gets more and more influential. The connection function for $l=1$ is depicted in Figure~\ref{fig:mesh_xor_1}. If the two \textbf{absolute} input values are different, the outcome of the $l=1$ variation gets closer to zero (somewhat similar to an AND connection). When the two inputs are close in their absolute value ($|x| \approx |y|$), the equation is essentially implementing a switch: positive outputs for two inverse inputs ($x \approx -y$), negative for two similar ones ($x \approx y$). 
    
    Unlike the $l=0$ version, the connection function itself has a considerable effect on the outcome, the input values are almost never left unchanged.
\end{itemize}

The two extremes realize opposing degrees of \textit{effectiveness}. The $l=0$ variation simply combines the two inputs like an OR function. For $l=0.5$ (see Figure~\ref{fig:mesh_xor_05}), we realize a middle ground with an XOR function, and the $l=1$ function resembles an AND connection (that also compares the signs of the two inputs) and alters the input values substantially.

For our experiments, we allow the network to learn the parameter $l$, using one separate scalar $l$ for each connected block. As described above, we assume that different layers of the network require different connection types. One could argue, for example, that early layers still contain a lot of redundant information that needs to be removed, while the process of forcefully removing or shifting information can hurt the accuracy in deeper layers. The learnable XOR connection is meant to address the issue by allowing the network to choose which configuration of the connection function to use for the current layer.

\section{Inserting Connected Blocks} 
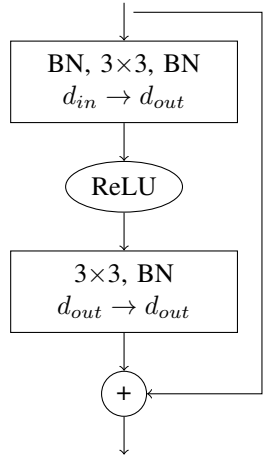
\begin{figure}
    \centering
        \begin{tikzpicture}[
            node distance=5mm and 5mm,
            block/.style={
                draw,
                minimum width={3cm},
            }]
            \node[] (b0) {};
            \node[block, below=of b0] (b1) {\begin{tabular}{c} BN, 3$\times$3, BN \\ $d_{in} \rightarrow d_{out}$\end{tabular}};
            \node[ellipse, draw, below=of b1] (b2) {ReLU};
            \node[block, below=of b2] (b3) {\begin{tabular}{c} 3$\times$3, BN \\ $d_{out} \rightarrow d_{out}$\end{tabular}};
            \node[circle, draw, below=of b3] (b4) {+};
            \node[below=of b4] (b5) {};

            \draw[->] (b0) edge (b1); 
            \draw[->] (b1) edge (b2); 
            \draw[->] (b2) edge (b3); 
            \draw[->] (b3) edge (b4); 
            \draw[->] (b4) edge (b5); 

            \draw [->] ($(b0.east) + (0, -0.25)$) -| +(1.7,-0.5) |- (b4.east);
        \end{tikzpicture}
    \caption{Structure of a basic block, illustrated as the sequence of actions performed and the number of input and output channels. 3$\times$3 indicates a convolution with kernel size 3, BN stands for batch normalization. The residual connection is visualized with the "+" node at the bottom. Like the connected block in Figure~\ref{fig:connected_block}, we use zero padding in the residual connection to make the block input match the block output.}
    \label{fig:basic_block}
\end{figure}

To evaluate the proposed connection functions, we chose an architecture closely resembling Min-nets \cite{gruning2022MinNets}. The network is comprised of two block types: the connected block, as described in the previous section, and a basic block, depicted in Figure~\ref{fig:basic_block}. The block with no connection is based on deep Pyramidal Residual Networks \cite{han2017pyramidResnet}: two convolutions with a 3$\times$3 filter, enhanced with batch normalization, ReLU activation, and a residual connection.

\begin{figure}
    \centering
        \begin{tikzpicture}[
            node distance=2mm and 10mm,
            block/.style={
                draw,
                minimum width={2cm},
                minimum height={1cm},
            }]
            \node[block] (b0) {\begin{tabular}{c}Stem\\$3\rightarrow16$\end{tabular}};  

            \node[block,below=of b0, fill={black!30}] (b11) {\begin{tabular}{c}Block 1-1\end{tabular}};  
            \node[block,below=of b11] (b12) {\begin{tabular}{c}Block 1-2\end{tabular}};  
            \node[block,below=of b12] (b13) {\begin{tabular}{c}Block 1-3\end{tabular}}; 

            \node[block,right=of b11, fill={black!30}] (b21) {\begin{tabular}{c}Block 2-1\\$16\rightarrow32$\end{tabular}};  
            \node[block,right=of b12] (b22) {\begin{tabular}{c}Block 2-2\end{tabular}};  
            \node[block,right=of b13] (b23) {\begin{tabular}{c}Block 2-3\end{tabular}}; 

            \node[block,right=of b21, fill={black!30}] (b31) {\begin{tabular}{c}Block 3-1\\$32\rightarrow64$\end{tabular}};  
            \node[block,right=of b22] (b32) {\begin{tabular}{c}Block 3-2\end{tabular}};  
            \node[block,right=of b23] (b33) {\begin{tabular}{c}Block 3-3\end{tabular}};  

            \draw [-, thick] (b13.south) |- +(1.5,-0.5) |- +(3, 4.1);
            \draw [->, thick] ($(b13.south) + (3, 4.1)$) -| (b21.north);

            \draw [-, thick] (b23.south) |- +(1.5,-0.5) |- +(3, 4.1);
            \draw [->, thick] ($(b23.south) + (3, 4.1)$) -| (b31.north);
        \end{tikzpicture}
    \caption{Structure of the network and connected block positions at the start of each stack, marked with a grey background.}
    \label{fig:network}
\end{figure}
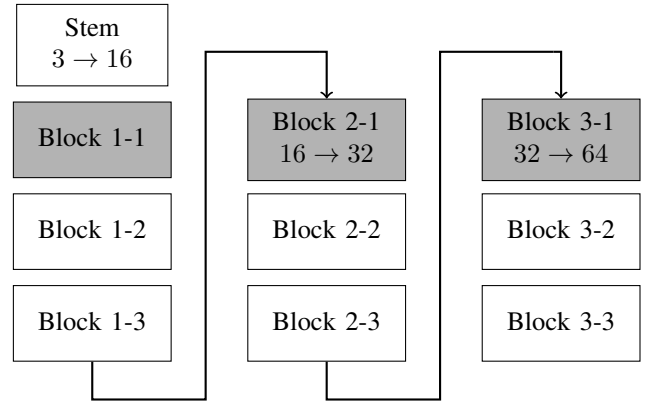

Our network is comprised of three stacks, each consisting of $N$ blocks, depicted in Figure~\ref{fig:network}. Every block is either a basic or a connected block.

As found in \cite{gruning2020FPYieldEffecientNetworks}, a network built of nothing but AND blocks is achieving lower accuracies than a network made up of basic blocks. Theoretically, a connected block could return the output of a basic block if needed. However, an AND connected block does not learn this behavior in practice \cite{gruning2020FPYieldEffecientNetworks}: replacing too many blocks of a network with an AND block decreases the accuracy. While the more flexible connection functions introduced in this paper are more likely to be capable of simulating a basic block, a network consisting of both basic and connected blocks always performed better than a network made up of only connected blocks.

Therefore, we insert connected blocks into specific positions. Our positioning is based on findings about a good positioning strategy in \cite{gruning2022FPInspiredByVision, gruning2022MinNets}: a connected block is employed at the beginning of each stack.

\section{Experiments} 
The insertion of AND blocks increases the accuracy of a CNN \cite{gruning2022FPInspiredByVision}. Can a different connection function increase the accuracy even further?  We evaluated several connection functions, to find that different connections are advantageous for different tasks. In this paper, we present the experimental results gained for the classification of Cifar10 \cite{krizhevsky2009cifar}, demonstrating that more general connection functions are capable of improving a network. 

The Cifar10 dataset features 60k 32x32 natural color images belonging to ten classes. The object defining the class usually covers large areas of the image, but is not always centered, sometimes partially occluded or seen from an unusual viewpoint, and sometimes identified by texture rather than shape. We use the standard data split: 50k images to train the model and the remaining 10k as validation set.

The experiments presented in this paper were done for a network with three blocks per stack (N=3), containing 217k trainable parameters. Similar to \cite{He2016ResNets}, we used momentum 0.9 and weight decay 0.0001, training for 200 epochs with stochastic gradient descent. The learning rate was initialized as 0.1 and multiplied by 0.1 after epochs 100 and 150. 

Each experiment was repeated for five different random seeds, taking note of the best accuracy observed over all epochs for each run. To gain a better insight into the development of the $l$ parameter (see Equation~\ref{eq:xor}), we separately performed 25 more repetitions of the Cifar10 first-per-stack experiment (resulting in a total of 30 repetitions).

\section{Results} 
\begin{figure}
    \centering
    \includegraphics[width=0.45\textwidth]{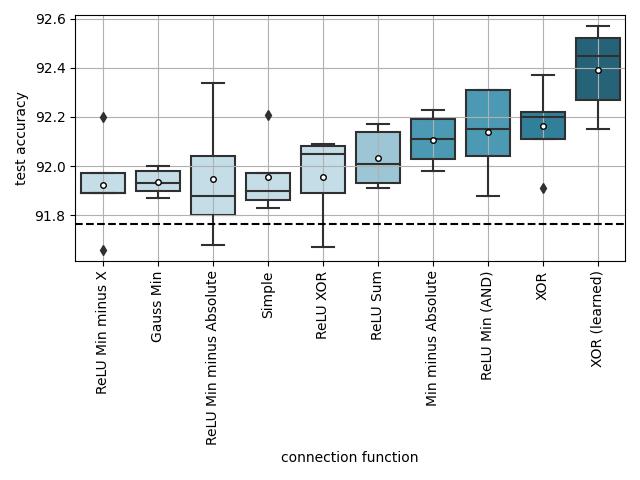}
    \caption{Test accuracy achieved using connected blocks with different connection functions, trained on Cifar10. The dashed line marks the accuracy of the baseline architecture with no connected blocks, the white dot in each box represents the mean value.}
    \label{fig:acc_cifar_fps}
\end{figure}

Our experiments reveal that a connected block generally increases the accuracy of the classification on Cifar10. 
All results depicted in Figure~\ref{fig:acc_cifar_fps} were obtained by the same underlying architecture (as presented in the previous sections), containing a total of 217k trainable parameters. 

To some extent, even the simple block (see Equation~\ref{eq:simple}) yields a higher test accuracy than the network with no connected block. Functions such as the ReLU min (i.e., the AND block), the "Min minus Absolute" (see Equation~\ref{eq:a_MMA} in the Appendix), or the XOR function (Equation~\ref{eq:xor}) further improve the accuracy. Thus, we can conclude that apart from an AND connection, other connection functions can perform just as well, or even slightly better.

However, one model achieves distinctly higher accuracies than all other networks: the learnable XOR connection. With a mean accuracy of 92.4\%, it outperforms the simple baseline block by 0.4\% and the baseline architecture with no connected block by 0.6\%. For comparison, a version of MobileNetV2 specifically modified to match the requirements of Cifar10 on a low parameter count achieved 92.4\% accuracy using 1.06M parameters \cite{Ayi2020ReducedMobilenet}, i.e., five times as many parameters as our network. We therefore assume that if the learnable parameter is applied effectively, the classification accuracy of the outcome is significantly increased. 

\begin{figure}
    \centering
    \includegraphics[width=0.4\textwidth]{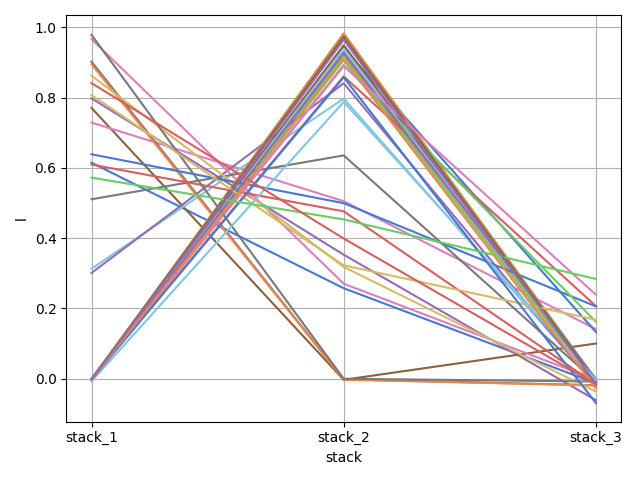}
    \caption{Learned parameter values for the XOR function for the three stacks of a first-per-stack network, trained using 30 different seeds on Cifar10. One line indicates the $l$ values learned during a single experiment, for the three connected blocks in stack one (earliest/topmost stack), two, and three (last stack).}
    \label{fig:xor_l}
\end{figure}

A closer look at the XOR parameters in 30 repetitions of the first-per-stack Cifar10 experiment reveals that the learned values follow the same pattern over all runs (see Figure \ref{fig:xor_l}): The parameter $l$ learned for the third stack ($l_3$) is always low. One of the two first stacks learns a higher value, up to $l\approx1$, while the other arrives at a smaller number, often zero. As described above, a small $l$ creates a connection that is not strictly transforming but mostly combining the two blocks. The fact that the last stack always learned this type of connection supports our assumption that more effective functions are best applied in earlier layers, and the ability to mimic a basic block becomes more useful in deeper layers. Allowing the network to choose different $l$ values (and therefore different variants of the XOR connection) in different stacks of the network makes the same connected block worthwhile on all positions. We assume that the learnable parameter allows the network to place AND ($l=1$) or OR ($l=0$) connections itself.

\section{Discussion} 
In this paper, we showed that CNNs could be improved by exchanging some of the basic convolution blocks with the proposed connection blocks, which employ two depthwise convolutions that are combined by specific connection functions. Although a simple summation of the filter values was already sufficient to beat the baseline, the proposed parameterized XOR connection function led to a significant further improvement. 

The learned parameter defines the degree to which the connection affects the outcome. Previous work hints at a drawback of the original AND connection: it sets large areas of the input to zero, which is useful in some layers, but possibly loses relevant data in others. The parameterized XOR is capable of realizing both a simple OR combination and a connection resembling the AND logic, allowing the network to choose the matching configuration for each layer.

While we improved upon the performance of small state-of-the-art networks on Cifar10, the main focus of our research is to provide a general \textit{proof of concept} for more diverse, parameterized connection functions used to combine the feature maps of two parallel convolutions. 

\section{Conclusion}

We have laid the groundwork for the future implementation of an even broader and more diverse set of connection functions, as we were able to demonstrate that (i) connected blocks can be an efficient method of increasing a network's accuracy and (ii) the training process is able to select a good option when given the choice between different variations of a connection function. 

The chosen function provides insights into the type of connection that works well in different positions. Our experiments indicate that the more restrictive AND connection works best in early layers as the network profits from filtering large parts of the redundant input at the beginning. Deeper networks will allow even more conclusions about the inner workings of the CNN. Adding parameterized connected blocks to a network allows the training process to adjust the structure to the task at hand, in a way that can be evaluated and understood.

Overall, we have demonstrated the potential of using pairs of neurons in the design of deep network architectures.



\bibliographystyle{IEEEtran}
\bibliography{IEEEabrv,bibliography.bib}

\appendix
The accuracy achieved by following connection functions is shown in Figure~\ref{fig:acc_cifar_fps}:

ReLU Min minus X (assymmetric Min minus Absolute: Only $x$ can produce a negative output):
\begin{equation}
    f(x, y) = \min(x^+, y^+) - 0.5 \cdot x^+,
    \label{eq:a_MMX}
\end{equation}

Gaus s Min (ReLU Min connection, but with smooth edges):
\begin{equation}
    f(x, y) = e ^{\frac{(x-y)^2}{50}},
    \label{eq:a_gaussmin}
\end{equation}

ReLU Min minus Absolute (positive when $x$ is close to $y$, negative if they are different, for $x, y \leq 0$ the function has a plateau: all inputs are set to zero):
\begin{equation}
    f(x, y) = \min(x^+, y^+) - 0.5 \cdot |x^+ - y^+|,
    \label{eq:a_reluMMA}
\end{equation}

Simple (both sides are merged without altering them):
\begin{equation}
    f(x, y) = (x + y)^+,
    \label{eq:a_simple}
\end{equation}

ReLU XOR (positive values only if either $x$ or $y$ is positive, while the other is negative of close to zero),
\begin{equation}
    f(x, y) = \max(x^+, y^+) - 0.5 \cdot (x^+ + y^+)
    \label{eq:a_reluXOR}
\end{equation}

ReLU Sum (return a value if either $x$ or $y$ is greater than zero, return zero otherwise):
\begin{equation}
    f(x, y) = x^+ + y^+,
    \label{eq:a_reluSUM}
\end{equation}

Min minus Absolute (Same as ReLU Min minus Absolute, but without the plateau: increasingly negative values when $x$ and $y$ are negative):
\begin{equation}
    f(x, y) = \min(x, y) - 0.5 \cdot |x - y|,
    \label{eq:a_MMA}
\end{equation}

ReLU Min (AND, return the smaller value only if both are greater than zero, return zero otherwise):
\begin{equation}
    f(x, y) = \min(x^+, y^+),
    \label{eq:a_min}
\end{equation}

XOR (XOR with no plateau, negative inputs now produce a negative result):
\begin{equation}
    f(x, y) = \max(x, y) - 0.5 \cdot (x + y)^+,
    \label{eq:a_XOR}
\end{equation}

XOR learned (XOR, with a learnable parameter $l$):
\begin{equation}
    f(x, y) = \max(x, y) - l \cdot (x + y)^+.
    \label{eq:a_learnedXOR}
\end{equation}

\end{document}